\documentclass[sort]{llncs}

\usepackage[T1]{fontenc}
\usepackage[utf8]{inputenc}
\usepackage[english]{babel}
\usepackage{cite}

\usepackage{multibib}
\newcites{appendix}{Additional sources}
\usepackage{filecontents}

\usepackage{amssymb}
\usepackage{mathtools}
\usepackage{hyperref}
\hypersetup{
  colorlinks   = true, %
  urlcolor     = blue, %
  linkcolor    = blue,  %
  citecolor    = blue   %
}
\usepackage{xspace}
\newcommand*{\eg}{e.g.\@\xspace}
\newcommand*{\ie}{i.e.\@\xspace}

\usepackage[export]{adjustbox}
\usepackage[nameinlink]{cleveref}

\newcommand{\bval}[1]{b = #1 s/mm$^2$}

\DeclareGraphicsExtensions{.png,.jpg}
\graphicspath{{images/}}

\makeatletter
\g@addto@macro\@floatboxreset\centering
\makeatother

\newcommand\blfootnote[1]{%
  \begingroup
  \renewcommand\thefootnote{}\footnote{#1}%
  \addtocounter{footnote}{-1}%
  \endgroup
}

\begin{document}

\mainmatter

\title{Automatic, fast and robust characterization of noise distributions for diffusion MRI}

\author{Samuel St-Jean \and Alberto De Luca \and Max A. Viergever \and Alexander Leemans \\ \email{samuel@isi.uu.nl}}
\institute{Center for Image Sciences, University Medical Center Utrecht, Utrecht, the Netherlands}

\maketitle
\blfootnote{Accepted in MICCAI 2018. The final authenticated version is available online at \url{https://doi.org/10.1007/978-3-030-00928-1_35}}

\begin{abstract}
Knowledge of the noise distribution in magnitude diffusion MRI images
is the centerpiece to quantify uncertainties arising from the acquisition process.
The use of parallel imaging methods, the number of receiver coils
and imaging filters applied by the scanner, amongst other factors, dictate the resulting signal distribution.
Accurate estimation beyond textbook Rician or noncentral chi distributions often requires information about the acquisition process
(\eg coils sensitivity maps or reconstruction coefficients), which is not usually available.
We introduce a new method where a change of variable naturally gives rise to a particular form of the gamma distribution for background signals.
The first moments and maximum likelihood estimators of this gamma distribution explicitly depend on the number of coils,
making it possible to estimate all unknown parameters using only the magnitude data.
A rejection step is used to make the method automatic and robust to artifacts.
Experiments on synthetic datasets show
that the proposed method can reliably estimate both the degrees of freedom and the standard deviation.
The worst case errors range from below 2\% (spatially uniform noise) to approximately 10\% (spatially variable noise).
Repeated acquisitions of \textit{in vivo} datasets show that the estimated parameters are stable and have lower variances than compared methods.
\end{abstract}

\section{Introduction}
\label{sec:intro}

Diffusion magnetic resonance imaging (dMRI) is a non invasive imaging technique which allows
probing the microstructure of the brain. Recent advances in parallel imaging techniques
and accelerated acquisitions have greatly reduced the inherently long scan time in dMRI.
While it is known that the noise distribution found in magnitude dMRI data depends
on the reconstruction algorithm used \cite{brown2014magnetic} and the number of channels in the
receiver coils\cite{Dietrich2008,Sotiropoulos2013b}, noise correlation effects in adjacent channels
change the noise distribution from its theoretical formulation.
Assumption of the Rician or more general noncentral chi distributions with degrees of freedom equal
to the number of receiver coils deviate due to these effects and other filtering applied by the scanner.
The resulting distribution usually exhibits a lower number of degrees of freedom $N$ than the number of receiver coils
and higher noise variance $\sigma_g$ depending on the spatial location\cite{Dietrich2008,Aja-Fernandez2014}.
Correcting deviations from the theoretical noise distributions is challenging and oftentimes requires
coils correlation maps or information about the complex signal combination process,
which is not readily available on most scanners.
While some recent algorithms for dMRI are developed to include information about the
noise distribution\cite{Collier2018,Sakaie2017}, there is no method, to the best of our knowledge,
providing a fully automatic way to characterize the noise distribution
using information from the magnitude data itself only.
Due to this gap between the physical acquisition process and noise estimation theory,
noise distributions are either assumed as Rician or noncentral chi with $N$ already known
and concentrate in estimating the noise standard deviation $\sigma_g$
\cite{Veraart2015a,Koay2009b,Tabelow2014}.
We propose to estimate both $\sigma_g$ and $N$ from the magnitude data only
by using a change of variable to a gamma distribution $\Gamma(N, 1)$ \cite{Koay2009b},
whose first moments directly depend on $N$.
This makes the proposed method fast and easy to apply on existing data without
additional information, while being robust to artifacts by only considering voxels
adhering to the created gamma distribution.

\section{Theory and Methods}
\label{sec:theory}
\paragraph{Signal distributions in parallel MRI}

To account for uncertainty in the acquisition process, the complex signal measured in k-space by the receiver coil
can be modeled with a separate additive zero mean Gaussian noise for each channel, but assumed to have identical variance $\sigma^2_g$.
When converted to the commonly used magnitude images, the resulting noise distribution follows a Rician or noncentral chi distribution,
whose parameters depend on the employed reconstruction algorithm \cite{Dietrich2008}.
To account for signal correlations introduced by parallel imaging techniques,
the case of the noncentral chi distribution is still valid with spatially varying parameters \cite{Aja-Fernandez2014}.

\paragraph{Parameter estimation using the method of moments and maximum likelihood}

When the underlying signal intensity $\eta$ is zero, the magnitude signal $m$ reduces to a Rayleigh distribution
or in the general case to a chi distribution.
The pdf of magnitude noise over zero signal is given by
$pdf(m | \eta = 0, \sigma_g, N) = (m^{2N-1}) / (2^{N-1}\sigma^{2N}_g \Gamma(N)) \exp{\left(-m^2 / (2\sigma_g^2)\right)} dm$
where $\Gamma(x)$ is the gamma function.
With the change of variable $t = m^2 / (2\sigma^2_g)$, the pdf can be rewritten as a gamma distribution $\Gamma(N, 1)$\cite{Koay2009b}.
The pdf of the gamma distribution $\Gamma(\alpha, \beta)$ is defined as
$pdf(t|\alpha,\beta) = 1 / (\Gamma(\alpha) \beta^\alpha) t^{\alpha-1} \exp{(-t / \beta)} dt$
and has theoretical mean $\mu_{gamma} = \alpha\beta$ and variance $\sigma^2_{gamma} = \alpha\beta^2$.
For a gamma distribution $\Gamma(N, 1)$, we obtain that the mean and the variance are equal with a value of $N$.
Another useful identity is that the sum of gamma distributions is a gamma distribution
such that if $t_i \thicksim \Gamma(\alpha_i, \beta)$,
then $\sum_{i=1}^K t_i \thicksim \Gamma(\sum_{i=1}^K\alpha_i, \beta)$.
We can therefore estimate the Gaussian noise standard deviation $\sigma_g$ and the number of coils $N$
from the moments of the magnitude image themselves where no signal from the imaged object is present.
Any method suitable for computing $\sigma_g$ can be used, or it can also be estimated from the moments once again with the relationship

\begin{equation}
    \sigma_g = \frac{1}{\sqrt{2}} \sqrt{\frac{\sum_{k=1}^K m^4_k}{\sum_{k=1}^K m^2_k} - \frac{1}{K}\sum_{k=1}^K m^2_k}
    \label{eq:find_sigma}
\end{equation}

where $m_k$ is the magnitude signal for voxel $k$ and $K$ is the number of identified noise only voxels.
Once $\sigma_g$ is known, $N$ can be estimated from the moments with

\begin{equation}
    N = \frac{1}{K}\sum_{k=1}^K t_k = \frac{1}{2K\sigma^2_g}\sum_{k=1}^K m^2_k
    \label{eq:moments_N}
\end{equation}

where $t_k = m_k^2 / (2\sigma_g^2)$ is the change of variable for voxel $k$.
Estimation based on the method of maximum likelihood yields two equations
for estimating $\alpha$ and $\beta$. Rearranging the equations for a gamma distribution $\Gamma(N, 1)$ \cite{Thom1958}
will give the same expression as \cref{eq:find_sigma} and a second implicit equation for $N$ that is given by

\begin{equation}
    \psi(N) = \frac{1}{K}\sum_{k=1}^K \log (m^2_k / 2\sigma_g^2)
    \label{eq:ml_N}
\end{equation}

where $\psi(x)$ is the digamma function and can be numerically inverted using Newton's method to obtain $N$.

\paragraph{Estimating $\sigma_g$ and $N$}

For simplicity, we assume that each 2D slice with the same spatial location belongs to the same distribution
throughout each 3D volume. This practical assumption allows selecting a large number of noise only voxels for computing statistics
as well as discarding acquisition artifacts such as ghosting.
Following a methodology similar to \cite{Koay2009b}, it is possible to identify voxels belonging to the
gamma distribution by checking if they fall inside a predefined probability threshold of the inverse cumulative distribution function (cdf).
Taking the sum of all MRI volumes can therefore be used to separate the background signal
belonging to the gamma distribution $\Gamma(KN, 1)$ from the rest of the volume with a rejection step using the inverse cdf.
In the particular case $\Gamma(\alpha, 1)$ at a probability level $p$,
the inverse cdf is $icdf(\alpha, p) = P^{-1}(\alpha, p)$ where $P^{-1}$ is the inverse lower incomplete regularized gamma function.
For the first iteration, initial bounds are set on the value of $N$ and $\sigma_g$ as they are unknown.
We set a lower bound $N_{min} = 1$ and an upper bound $N_{max} = 12$ for the first iteration, noting that \cite{Dietrich2008}
reported values of $N$ between 3 and 12 for a 32 channels receiver coil.
Similar to \cite{Koay2009b},
an upper bound of $\sigma_g$ is given by $\sigma_{g_{max}} = median / \sqrt{2\, icdf(N_{max}, 1/2)}$
where $median$ is the median of the whole 4D dMRI dataset.
From this upper bound $\sigma_{g_{max}}$, a search interval with $a$ values is created, where we chose $a = 50$ as in \cite{Koay2009b}.
Each point of the interval $\Phi = [1 \sigma_{g_{max}} / a, 2 \sigma_{g_{max}} / a, \dotsc , a \sigma_{g_{max}} / a]$ is used
as an initial value of $\sigma_g$ in the change of variable $t = m^2 / 2\sigma_g^2$.
With these initial values, an iterative search for $\sigma_g$ and $N$ is made as follow.
The value of $\Phi$ which identifies the largest number of voxels
between the lower bound given by $\lambda_{-} = icdf(KN_{min}, p/2)$ and the upper bound given by $\lambda_{+} = icdf(KN_{max}, 1-p/2)$
is accepted as $\sigma_g$.
From those voxels, new values of $\sigma_g$ are computed with \cref{eq:find_sigma} and $N$ with \cref{eq:moments_N} or \cref{eq:ml_N}.
For the next iteration, we set $\Phi = [0.95 \sigma_{g}, 0.96 \sigma_{g}, \dotsc , 1.05 \sigma_{g}]$ and
recompute the $icdf$ bounds $\lambda_{-}, \lambda_{+}$ with the new value of $N$.
Voxels between $\lambda_{-}$ and $\lambda_{+}$ belong to the distribution $\Gamma(KN, 1)$
and are recomputed until the values of $\sigma_g$ and $N$ reach convergence.

\paragraph{Synthetic phantom datasets}

We generated synthetic datasets based on the ISBI 2013 HARDI
challenge\footnote{\url{http://hardi.epfl.ch/static/events/2013_ISBI/}} with
phantomas\footnote{\url{https://github.com/ecaruyer/phantomas}}.
Two noiseless single shell phantoms with 64 gradient directions
were generated at \bval{1000} and \bval{3000} with one \bval{0} each.
The datasets were then corrupted with Rician $(N = 1)$ and noncentral chi noise ($N = $ 4, 8 and 12),
both stationary and spatially varying, at a signal-to-noise ratio (SNR) of 30.
The noisy data was generated according to
$\hat{I} = \sqrt{\sum_{i=0, j=0}^{N} \left(\frac{I}{\sqrt{N}} + \tau\epsilon_i \right)^2 + \tau\epsilon_j^2}$,
where $\hat{I}$ is the resulting noisy volume $\epsilon_i, \epsilon_j$ are Gaussian distributed with mean 0 and
variance $\sigma_g^2 = (mean(b0)/SNR)^2$.
In the constant noise case, $\tau$ is set to 1 so that the noise is uniform.
For the spatially varying noise case, $\tau$ is a sphere with a value of 1 in the center up to a value of 1.75 at the edges of the phantom,
thus generating a stronger noise profile \textit{outside} the phantom than for the stationary (constant) noise case.
This noise profile mimics
receiver coils disposed around the surface of the phantom, with an increase in the noise profile near each receiver.
One important observation arising from choosing a single SNR level
is that the noise standard deviation $\sigma_g$ is the same for all datasets, while the
magnitude standard deviation $\sigma_{m_N}$ depends on the value of $N$ and we have $\sigma_{m_N} < \sigma_g$.

\paragraph{\textit{In vivo} datasets}

We obtained four repetitions of a freely available dMRI dataset of a single
subject\footnote{\url{https://openfmri.org/dataset/ds000031}} to assess the reproducibility
of noise estimation without \textit{a priori} knowledge.
The acquisition was performed on a GE MR750 3T scanner at Stanford university,
where a 3x slice acceleration with blipped-CAIPI shift of FOV/3 was used, partial Fourier 5/8 and a minimum TE of 81 ms.
Two acquisitions were made in the anterior-posterior phase encoded direction and the two others
in the posterior-anterior direction.
The voxelsize was 1.7 mm isotropic with 7 \bval{0} images, 38 volumes at \bval{1500} and 38 volumes at \bval{3000}.

\paragraph{Noise estimation algorithms for comparison}

To assess the performance of the proposed method, we used three other noise estimation algorithms\cite{Veraart2015a,Koay2009b,Tabelow2014}
previously used in the context of diffusion MRI with their default parameters.
The local adaptive noise estimation (LANE) algorithm \cite{Tabelow2014} was designed
to estimate the noise standard deviation over tissue for both Rician and noncentral chi
noise while also taking into account the structure of the data for adaptive estimation.
Since the method works on a 3D volume, we only used the \bval{0} image for all of the experiments
as the signal does not vary spatially for the same type of tissue in such image.
We used the Marchenko-Pastur (MP) distribution fitting on the principal component analysis (PCA) decomposition of the diffusion data\cite{Veraart2015a}.
MPPCA estimates the magnitude noise standard deviation $\sigma_{m_N}$ in small local windows by finding an optimal threshold in PCA space
which separates the signal from the noise.
This value of $\sigma_{m_N}$ is slightly underestimated due to the discrete nature of the PCA decomposition.
Finally, we compared our proposed method with the Probabilistic Identification and Estimation of Noise (PIESNO) \cite{Koay2009b},
which originally proposed the change of variable to the gamma distribution that is at the core of our proposed method.
PIESNO requires the value of $N$, which is used to iteratively estimate $\sigma_g$ until convergence
by removing voxels which do not belong to the distribution $\Gamma(N, 1)$ for a given slice.
For our proposed algorithm, we set the probability level at $p = 0.05$ and initial values of $a = 50$, $N_{min} = 1$ and $N_{max} = 12$.
To the best of our knowledge, ours is the first method which makes it possible to estimate both
$\sigma_g$ and $N$ jointly without requiring any information about the reconstruction process of the MRI scanner.
Finally, we quantitatively assessed the performance of each method on the synthetic datasets by measuring the
percentage error inside the phantom against the known value of $\sigma_g$, where the error is computed as
$percentage\, error = 100 (\sigma_{g_{estimated}} - \sigma_{g_{true}}) / \sigma_{g_{true}}$.
We also show the estimated values of $N$ using our method for each dataset.

\section{Results}

\begin{figure}[ht]
    \includegraphics[width=0.75\textwidth]{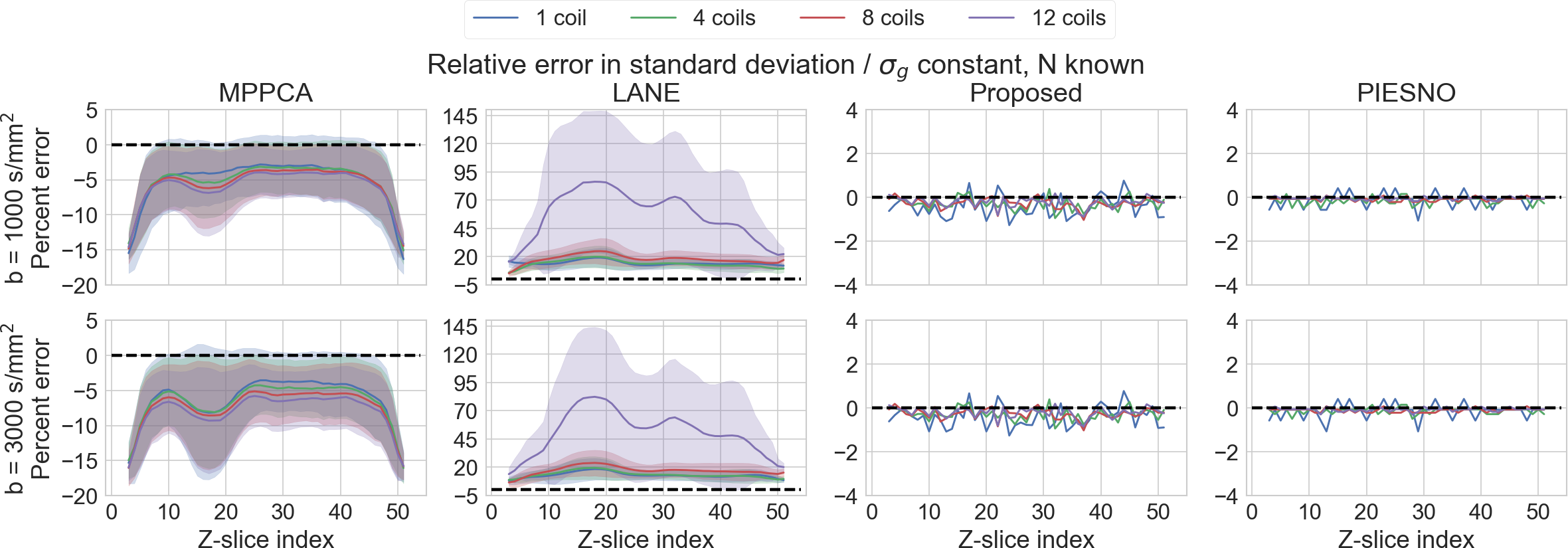}
    \includegraphics[width=0.75\textwidth]{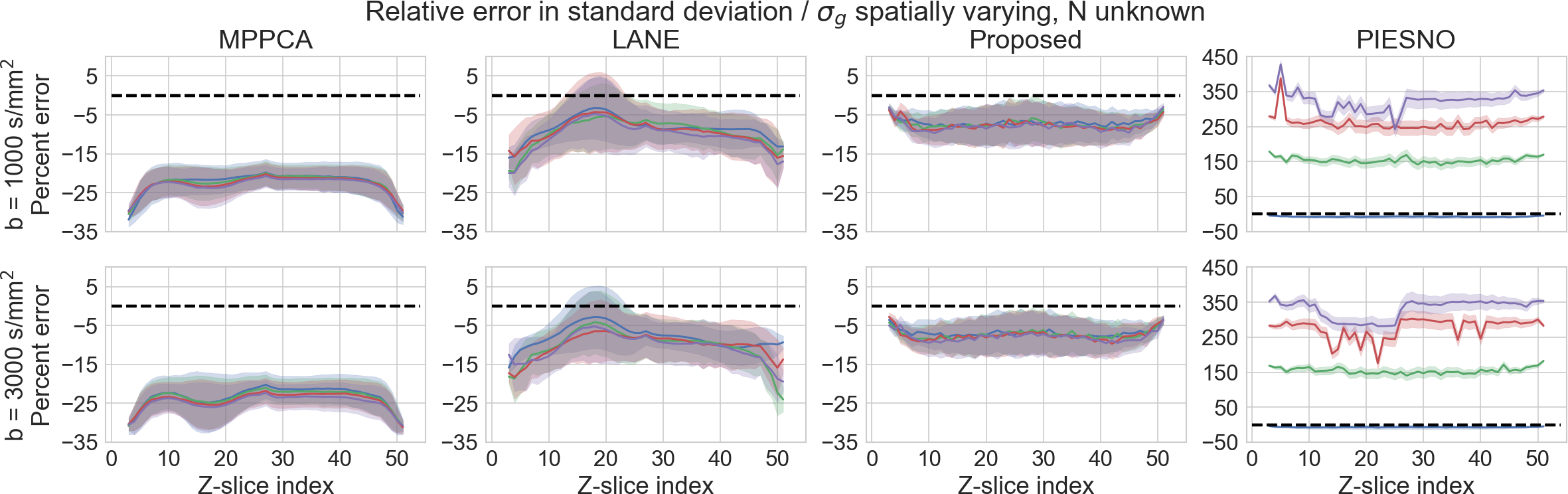}
    \caption{Percentage of error in estimating the noise standard deviation for each slice along the Z axis with the mean (solid line) and standard deviation (shaded area).
    In the top image, $\sigma_g = 171$ is constant and $N$ is known
    while in the bottom image $\sigma_g$ varies spatially and $N$ is unknown or assumed Rician distributed.}
    \label{fig:phantomas_sigma}
\end{figure}

\begin{figure}[ht]
    \includegraphics[width=0.75\textwidth]{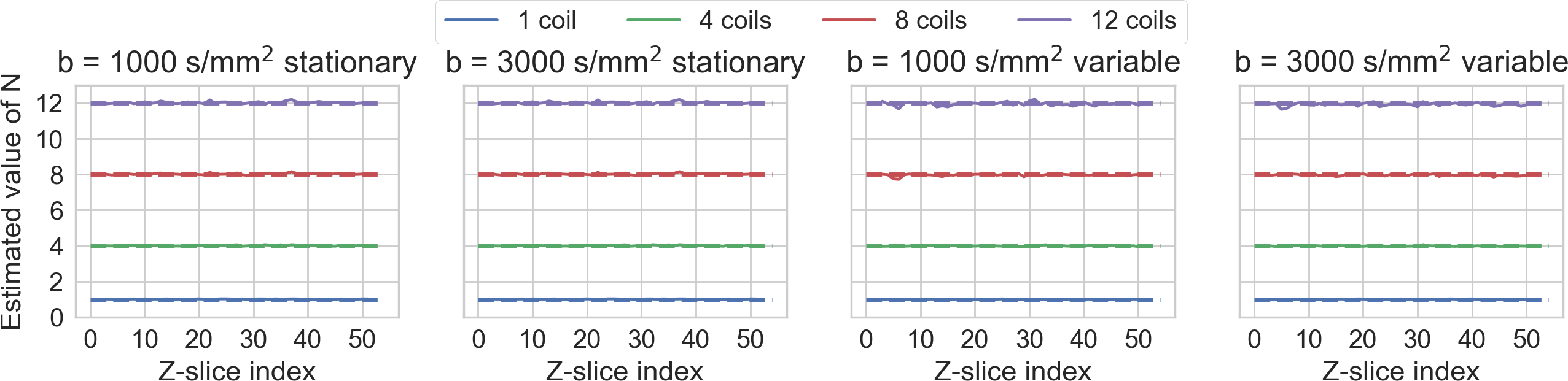}
    \caption{Estimated value of $N$ by the proposed method. Even for the spatially variable case where $\sigma_g$ is slightly underestimated,
    the estimated values of $N$ are stable and correspond to the real values used in the synthetic simulations in every case.}
    \label{fig:phantomas_estimating_N}
\end{figure}

\begin{figure}[ht]
    \includegraphics[width=0.75\textwidth]{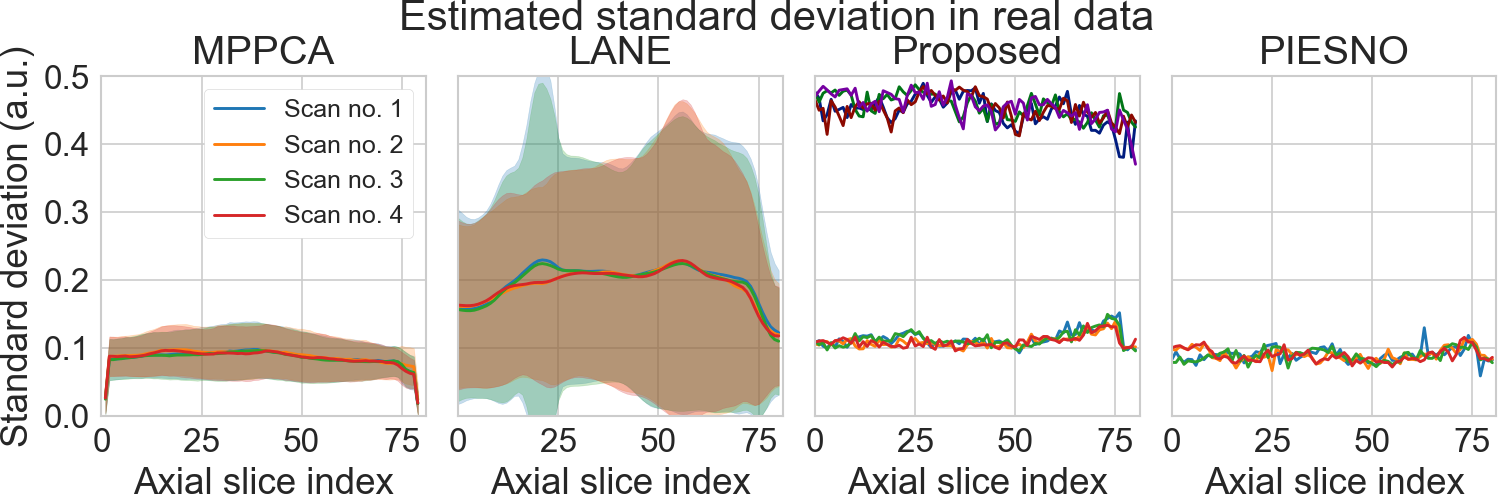}
    \includegraphics[width=0.75\textwidth]{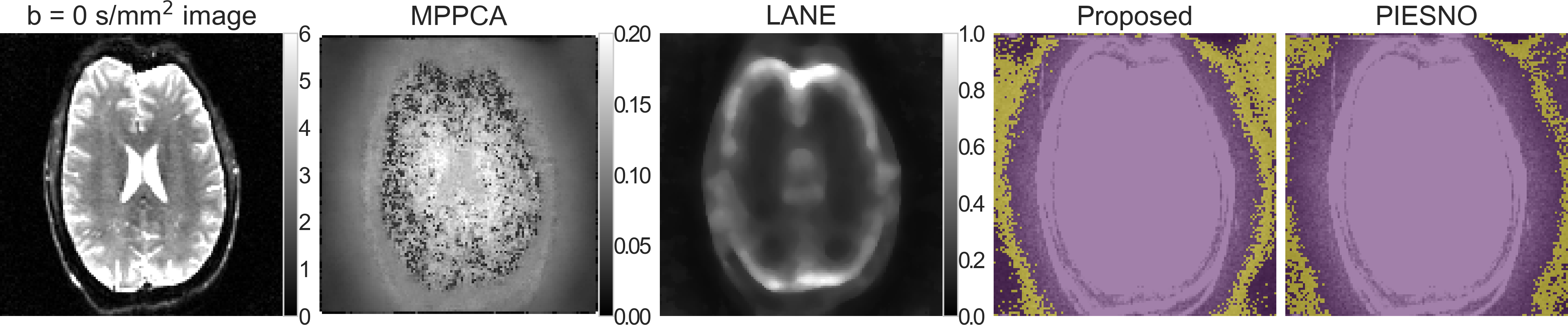}
    \caption{At the top, estimated values of $\sigma_g$ for the 4 \textit{in vivo} datasets.
    For the proposed method, estimated values of $N$ are shown in darker hues for each dataset.
    On the bottom, an axial slice of a \bval{0} image from one dataset and the estimated values of $\sigma_g$ for MPPCA and LANE.
    For the proposed method and PIESNO, a mask of the identified background voxels (in yellow) overlaid on the data.}
    \label{fig:openfmri_std}
\end{figure}

\begin{figure}[!ht]
    \includegraphics[valign=t,width=0.65\textwidth]{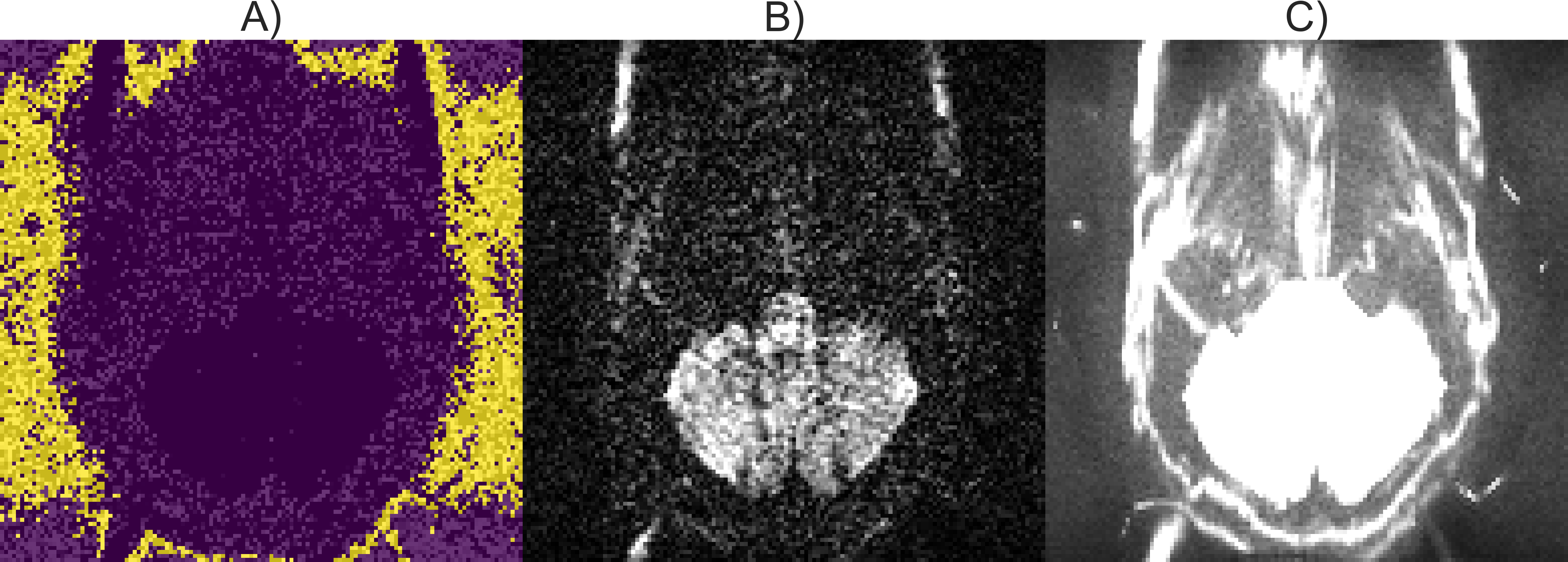}
    \hfill
    \includegraphics[valign=t,width=0.34\textwidth]{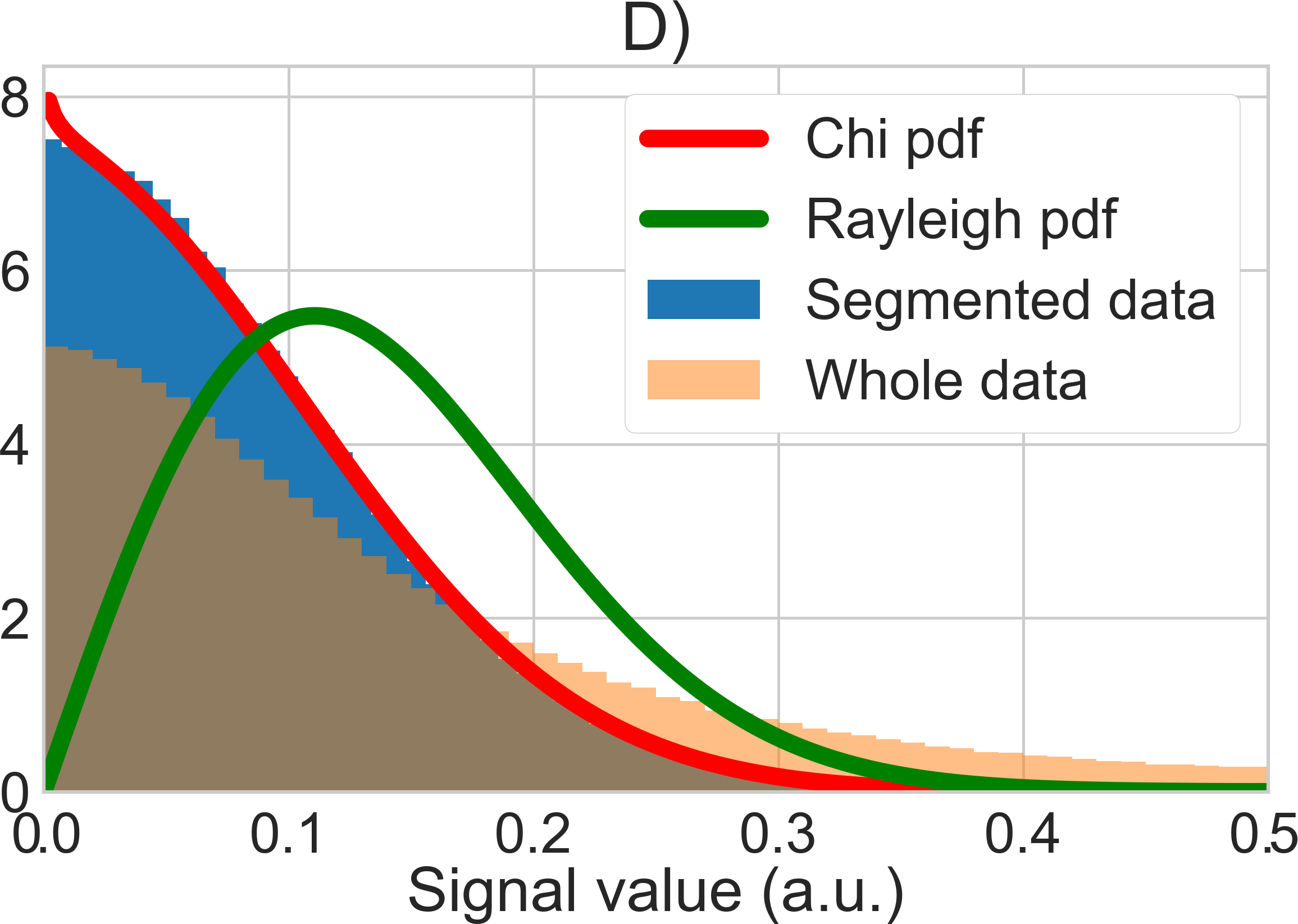}
    \caption{An axial slice in the cerebellum from one of the \textit{in vivo} datasets.
    Voxels identified in A) as noise only (yellow) are free of artifacts in a single slice in B) or along the sum of all volumes in C).
    In D), the normalized density histogram using the selected voxels from A) (blue)
    fits well a chi distribution with $N = 0.47$ and $\sigma_g = 0.11$, while assuming a Rayleigh distribution (green)
    or using all non brain voxels (orange) leads to a worse visual fit.}
    \label{fig:openfmri_mask}
\end{figure}

As MPPCA and LANE are designed to estimate $\sigma_g$ over data, we report the estimation error
computed only inside a mask excluding the background for all methods.
\Cref{fig:phantomas_sigma} shows the percentage error of each method on the synthetic datasets.
The correct value of $N$ was given to both LANE and PIESNO when $\sigma_g$ was constant.
All methods performed generally well, with our proposed method and PIESNO making less than 2\% of errors for all cases.
MPPCA and LANE commit larger errors (around 5\% and 20\% on average respectively) with increasing values of $N$, where LANE
error is the largest when $N = 12$.
For the case of spatially varying $\sigma_g$, we assumed $N$ to be unknown and set $N = 1$ for LANE and PIESNO.
Due to a misspecification of $N$, PIESNO errors are several orders of magnitude larger than the other methods except for the Rician noise case.
MPPCA and LANE both underestimate $\sigma_g$ (around 20\% and between 10 to 15\% respectively) while our proposed method resulted in the lowest error, which is around 10\%.
\Cref{fig:phantomas_estimating_N} shows the estimated values of $N$ by the proposed method for all cases of
the synthetic datasets.
Even when $\sigma_g$ is underestimated, values of $N$ are close to the real value.
Estimating $N$ using \cref{eq:moments_N} or \cref{eq:ml_N} gave similar results in both cases, so we used \cref{eq:moments_N} in the present work.
As limited information is available for the \textit{in vivo} datasets, we assumed a Rician distribution for LANE and set $N = 1$
as suggested by \cite{Tabelow2014}.
For PIESNO, setting a Rician distribution with $N = 1$ returned less than 10 voxels identified per datasets.
We instead assumed $N = 0.5$ since it corresponds to a half Gaussian distribution \cite{Dietrich2008}, which is the closest theoretical distribution
estimated by our method.
\Cref{fig:openfmri_std} shows the mean (and standard deviation) value of $\sigma_g$
on the \textit{in vivo} datasets for each methods along axial slices.
The value of $N$ as computed by our proposed method is also reported and is stable across datasets.
All methods recovered average stable values of $\sigma_g$ on the four repetitions of the same subject.
However, LANE recovered the highest values of $\sigma_g$ amongst all methods with a large variance, which might indicate overestimation in some areas.
\Cref{fig:openfmri_mask} shows an axial slice around the cerebellum corrupted by acquisition artifacts likely due to parallel imaging.
Voxels containing artifacts were automatically discarded by our method. The values of $N$ and $\sigma_g$ computed from these voxels also offer
a better qualitative fit than assuming a Rayleigh distribution or selecting non brain data.
We also timed each method to estimate $\sigma_g$ on one of the \textit{in vivo} datasets using a standard desktop computer with a 3.5 GHz Intel Xeon processor.
All methods were multi threaded while PIESNO was only single threaded.
The runtime to estimate $\sigma_g$ (and N) was around 10 secs for our proposed method, 11 secs for PIESNO,
3 mins for MPPCA and 18 mins for LANE.

\section{Discussion and Conclusion}

We have shown how a change of variable to a gamma distribution $\Gamma(N, 1)$ can be used to robustly and automatically identify background voxels.
Once identified, the moments and maximum likelihood equations (\cref{eq:ml_N,eq:moments_N,eq:find_sigma}) of the gamma distribution can
be used iteratively to compute the number of degrees of freedom $N$ and the Gaussian noise standard deviation $\sigma_g$ relating to the original noise distribution.
The presented equations are also fast to compute (around 10 seconds on \textit{in vivo} data).
Results on the synthetic datasets show that we can reliably estimate both parameters from the magnitude data itself.
While the method we have presented assumes that each 2D slice contains a single noise distribution,
$N$ can be computed reliably on spatially varying noise and $\sigma_g$ with an error between 5 and 10\%, which is less than the compared methods.
On the \textit{in vivo} datasets, our method is stable across the four repetitions and can automatically discard voxels corrupted by acquisition artifacts due to parallel acceleration.
From the identified background voxels, without any specific assumption, the recovered distribution parameters fit well the histogram of the data.
This distribution is close to a half Gaussian distribution ($N = 0.5$) while the Rician noise assumption would not be adequate in this case.
Our method is also the first to identify any type of noise distribution from the magnitude data itself without requiring external information about the scanner or the reconstruction process.
Interestingly, while we have shown results on dMRI datasets, the theory we presented applies to any other MRI weighting using large samples of magnitude data \eg functional MRI.
If measurements from the scanner without any object signals are acquired (\ie noise maps), a local window estimation of our proposed method could be used to overcome the shortcoming
of assuming stationary 2D noise distributions.
Noise maps measurements could also be used for cases such as body or cardiac imaging where background voxels are usually not available in large quantities.
Automatic identification of the noise distribution parameters could help multicenter studies which may not currently collect
information about the acquisition and reconstruction process \cite{Aja-Fernandez2014}
or methods harmonizing data between different scanners and acquisition protocols \cite{Mirzaalian2016}.
Our method can also be used to provide prior knowledge beyond the textbook Rician distribution when computing local diffusion models \cite{Sakaie2017,Collier2018}.

\bibliographystyle{splncs}
\bibliography{stjean_etal_miccai_2018.bbl}

\section*{Acknowledgments}

We would like to thank Félix Camirand and Chantal M. W. Tax for ideas and discussions.
We additionally thank Maxime Descoteaux, Mariëlle Jansen, Michael J. van Rijssel and Majd Zreik for comments and general help in improving the manuscript.
Samuel St-Jean is supported by the Fonds de recherche du Québec – Nature et technologies (FRQNT).
This research is supported by VIDI Grant 639.072.411 from the Netherlands Organization for Scientific Research (NWO).

\appendix
\renewcommand{\theequation}{\Alph{chapter}.\arabic{equation}}
\section{Proofs of the main equations}
\subsection{Proof of \Cref{eq:find_sigma}}

The first two moments of the Gamma distribution  $\Gamma(\alpha, \beta)$ are given by

\begin{gather}
    \mu_{gamma} = \alpha\beta,\,
    \sigma^2_{gamma} = \alpha\beta^2
    \label{eq:moments_gamma}
\end{gather}

For the special case $\Gamma(N, 1)$, we therefore have

\begin{equation}
    \mu_{gamma} = N = \sigma^2_{gamma}
\end{equation}

Which we can compute using the sample mean and sample variance formulas such that

\begin{equation}
  N = \frac{1}{K}\sum_{k=1}^K t_k = \frac{1}{K}\sum_{k=1}^K t_k^2 - \left(\frac{1}{K}\sum_{k=1}^K t_k\right)^2
\end{equation}

Substituting the equation for the sample moments in terms of $t=\frac{m^2}{2\sigma^2}$, we obtain

\begin{align}
    & \frac{1}{K}\sum_{k=1}^K \frac{m^2_k}{2\sigma^2} = \frac{1}{K}\sum_{k=1}^K \left(\frac{m^2_k}{2\sigma^2}\right)^2 - \left(\frac{1}{K}\sum_{k=1}^K \frac{m^2_k}{2\sigma^2}\right)^2 \\
    \Rightarrow \quad & \frac{1}{2K\sigma^2}\sum_{k=1}^K m^2_k = \frac{1}{4K\sigma^4}\sum_{k=1}^K m^4_k - \frac{1}{4K^2\sigma^4}\left(\sum_{k=1}^K m^2_k\right)^2 \\
    \Rightarrow \quad & \sum_{k=1}^K m^2_k = \frac{1}{2K\sigma^2}\left(K\sum_{k=1}^K m^4_k - \left(\sum_{k=1}^K m^2_k\right)^2\right) \\
    \Rightarrow \quad & 2K\sigma^2 = \frac{K \sum_{k=1}^K m^4_k - \left(\sum_{k=1}^K m^2_k\right)^2}{\sum_{k=1}^K m^2_k} \\
    \Rightarrow \quad & \sigma = \frac{1}{\sqrt{2K}} \sqrt{\frac{K \sum_{k=1}^K m^4_k - \left(\sum_{k=1}^K m^2_k\right)^2}{\sum_{k=1}^K m^2_k}}\\
    \Rightarrow \quad & \sigma = \frac{1}{\sqrt{2}} \sqrt{\frac{\sum_{k=1}^K m^4_k}{\sum_{k=1}^K m^2_k} - \frac{1}{K}\sum_{k=1}^K m^2_k}
\end{align}

which is the form of \cref{eq:find_sigma}.

\subsection{Proof of \Cref{eq:moments_N}}

With the variance $\sigma^2$ now known, \cref{eq:moments_gamma} yields \cref{eq:moments_N} directly

\begin{equation}
    N = \mu_{gamma} = \frac{1}{2K\sigma^2}\sum_{k=1}^K m^2_k
    \label{eq:gamma_moment_N}
\end{equation}

\subsection{Proof of \Cref{eq:ml_N}}

For the Gamma distribution $\Gamma(\alpha, \beta)$, maximizing the log likelihood by equating the partial derivative to 0 for each parameter will give the two equations \cite{Thom1958}

\begin{align}
    \frac{1}{K\beta}\sum_{k=1}^K t_k - \alpha &= 0
    \label{eq:gamma_ml1}\\
    \log(\beta) + \frac{\partial}{\partial\alpha} \log(\Gamma(\alpha)) - \frac{1}{K}\sum_{k=1}^K \log(t_k) &= 0
    \label{eq:gamma_ml2}
\end{align}

Since we have $\alpha = N$ and $\beta = 1$, in this special case \cref{eq:gamma_ml1} is the same as \cref{eq:gamma_moment_N}
and \cref{eq:gamma_ml2} can be rewritten as an implicit equation of $N$

\begin{align}
    \psi(N) &= \frac{1}{K}\sum_{k=1}^K \log (m^2_k / 2\sigma^2) \\
    \Rightarrow \quad N &= \psi^{-1}\left(\frac{1}{K}\sum_{k=1}^K \log (m^2_k / 2\sigma^2)\right)
    \label{eq:gamma_ml_N}
\end{align}

where $\psi(x) = \frac{d}{dx}\log(\Gamma(x))$ is the digamma function.
\Cref{eq:gamma_ml_N} can be solved numerically using Newton's method to find $x = \psi^{-1}(y)$.
A starting estimate for $x_0$ is given by \citeappendix{Minka2003}

\begin{displaymath}
    x_0 = \psi^{-1}(y)\approx
    \begin{dcases}
        \exp(y)+ 1/2        & \text{if } y \geq -2.22\\
        -1 / (y + \psi(1))  & \text{if } y <    -2.22\\
    \end{dcases}
\end{displaymath}

The update rule for Newton's method at iteration $n$ is therefore

\begin{equation}
    x_{n+1} = x_{n} - \frac{\psi(x_{n}) - y}{\psi'(x_{n})}
    \label{eq:newton}
\end{equation}

where $\psi'$ is the derivative of $\psi$ and is called the polygamma function.


\bibliographystyleappendix{splncs}
\bibliographyappendix{appendix}

\end{document}